\documentclass[a4paper]{article}

\usepackage[pages=all, color=black, position={current page.south}, placement=bottom, scale=1, opacity=1, vshift=5mm, contents={DRAFT}]{background}

\usepackage[margin=1in]{geometry} 

\usepackage[sort&compress,numbers,square]{natbib}
\usepackage{subfigure}
\usepackage{booktabs}
\usepackage{multirow}
\usepackage{hyperref}
\usepackage{amsmath}

\title{AI Act Evaluation Benchmark: An Open, Transparent, and Reproducible Evaluation Dataset for NLP and RAG Systems}
\author{Athanasios Davvetas$^1$ \and Michael Papademas$^1$\textsuperscript{,}$^2$ \and Xenia Ziouvelou$^1$ \and Vangelis Karkaletsis$^1$}

\date{
	$^1$National Centre for Scientific Research ``Demokritos'', Institute of Informatics and Telecommunications, Aghia Paraskevi, Greece \\ \texttt{\{tdavvetas, papademasmichael, xeniaziouvelou, vangelis\}@iit.demokritos.gr} \\ 
    \vspace{0.5em}
    $^2$Department of Communication, Media and Culture, Panteion, University of Social and Political Sciences, Athens, Greece
}

\begin{document}
\maketitle

\begin{abstract}
    The rapid rollout of AI in heterogeneous public and societal sectors has subsequently escalated the need for compliance with regulatory standards and frameworks. The EU AI Act has emerged as a landmark in the regulatory landscape. The development of solutions that elicit the level of AI systems' compliance with such standards is often limited by the lack of resources, hindering the semi-automated or automated evaluation of their performance. This generates the need for manual work, which is often error-prone, resource-limited or limited to cases not clearly described by the regulation. This paper presents an open, transparent, and reproducible method of creating a resource that facilitates the evaluation of NLP models with a strong focus on RAG systems. We have developed a dataset that contain the tasks of risk-level classification, article retrieval, obligation generation, and question-answering for the EU AI Act. The dataset files are in a machine-to-machine appropriate format. To generate the files, we utilise domain knowledge as an exegetical basis, combining with the processing and reasoning power of large language models to generate scenarios along with the respective tasks. Our methodology demonstrates a way to harness language models for grounded generation with high document relevancy. Besides, we overcome limitations such as navigating the decision boundaries of risk-levels that are not explicitly defined within the EU AI Act, such as limited and minimal cases. Finally, we demonstrate our dataset's effectiveness by evaluating a RAG-based solution that reaches 0.87 and 0.85 F1-score for prohibited and high-risk scenarios. 

    \noindent\textbf{Keywords:} EU AI Act, NLP models, RAG evaluation, Large Language Models, Raw Data Processing, Label Scarcity
\end{abstract}

\section{Introduction}
The technological advancements in the field of Artificial Intelligence (AI) have enabled its access and applications to a variety of scientific subjects. Some examples include applications in medicine \cite{Pozdeyev2025},\cite{Shariatnia2025}, construction management \cite{Zhang2024} and even reaching the status of general purpose technology \cite{Crafts2021}. However, its development and application has initiated a pursuit of establishing, maintaining, and complying with regulation standards. As highlighted in \cite{Perboli2025}, the embrace of AI systems creates ethical, societal, and regulatory challenges. The EU AI Act is a landmark in the AI regulatory landscape which safeguards the operation of these systems in a way that balances innovation with regulatory compliance. However, their evolvement into general-purpose technologies has created even more challenges which indicate a shift from reactive to proactive AI governance \cite{Gstrein2024}. Despite, their general-purpose status is challenging not only to the broader scope of the regulations but also to their consistent application \cite{Taeihagh2025}. 

However, regulatory compliance is a process that requires a diverse set of expertise, and is typically manual, necessitating the call for solutions that enable automated or semi-automated compliance \cite{Kulkarni2021}. Manual work is costly in terms of resources, as it requires significant amounts of effort, or may be limited due to requiring a very particular set of skills (e.g., legal expertise). Besides, it is frequent prone to errors and hard to maintain with further changes or refinements in the regulatory landscape. The European and global regulatory landscape is constantly adapting to technological advancements, which increasingly highlights the importance of automated or semi-automated solutions to address the above challenges.

The rise of Retrieval-Augmented Generation (RAG) systems and other relevant natural language processing (NLP) models has highlighted the extraction of structured data from PDFs as challenging \cite{Atagong2025}. Extracting regulatory text is no exception. The lack of structured data has a large impact on the standardisation of evaluating NLP systems, while also making machine-to-machine access cumbersome due to PDFs being optimised for visualisation, rather than structure. In turn practitioners either have to put in manual work to convert them to machine-readable formats, or rely on unofficial processed datasets released in various platforms, such as HuggingFace or Github. Such datasets may not be properly documented which affects their transparency and document relevance. Not being able to employ automated evaluations hinders the development and deployment of such solutions or renders them less accessible, not transparent or error-prone.

In this paper, we propose an open, reproducible and transparent method to extract a dataset for the evaluation of AI models that facilitate the compliance with the EU AI Act. Our focus is on the evaluation of NLP models, with a focal point on RAG systems. Our approach promotes the holistic automated evaluation of such systems, by involving multiple facets of machine learning tasks extracted from the regulations. Tasks such as risk-level classification, relevant article extraction, obligation generation and question answering (QA), enables a comprehensive benchmarking of models that aim to support the compliance with the regulation. Our approach leverages domain knowledge extracted directly from raw text along with the processing and reasoning power of modern large language models (LLMs) in an attempt to reduce manual work, standardise output and create easy to use machine-readable output format (JSON), while simultaneously providing document-grounded generation.

Our work actively contributes to the development of automated solutions that facilitate compliance with the EU AI Act, by providing coverage of machine learning tasks extracted from the regulation itself. First, we generate a set of scenarios in order to create a statistical decision boundary for risk-level classification. Besides, this task enables the development of AI systems for rapid risk-level assessment and acts as a quantifiable metric for the reasoning capabilities of an NLP system. The task of extracting relevant articles support the testing retrieval, while actionable steps to compliance or obligations test its generation capabilities. Finally, QA pairs facilitate the evaluation of agentic properties. This approach covers a variety of desired properties for NLP systems operating on the EU AI Act, while at the same time facilitating the democratisation of developing such systems for the purpose of regulation compliance. However, several other uses of the dataset can be identified beyond practitioners, such as educational purposes or refinement of skills and knowledge from legal experts.

The rest of the paper is organised as follows: Section \ref{sec:relwork} presents the related work, Section \ref{sec:methodology} details our methodology for the production of the dataset and background knowledge required, Section \ref{sec:overview} presents an overview of the produced dataset, while Section \ref{sec:use-case} presents a use-case of utilising the produced dataset for the task of risk-level classification. Finally, Section \ref{sec:conc} details conclusions, limitations and future directions. 

\section{Related Work}
\label{sec:relwork}
In this section, we explore the current landscape on data generation for the evaluation of AI models in the context of regulatory compliance. To the best of our knowledge, related work to that particular field is rather limited, but as mentioned in the previous section, it is currently an emerging trend that has been increasingly attracting attention from the scientific community. Besides the scientific community, attention from the side of policy-maker's, SMEs, developers, legal experts, and other similar stakeholders on the side of AI development and deployment is emerging. Therefore, attention has been drawn in creating frameworks that standardise and categorise AI systems. The OECD classification for AI systems \cite{OECD2022} is one of the foundational approaches to establishing a resource to aid policymakers, regulators and legislators, to categorise and evaluate AI systems from a policy perspective. While an important stepping stone, the OECD approach is designed for manual assessment and evaluation by experts, and thus requiring manual effort to translate into automated evaluation or benchmarks. 

The process of dataset curation and generation, especially for training purposes has been gaining more and more attention. Such studies mainly focus into ethical practices, such as fairness or privacy, among others. However, regulatory requirements and compliance with such regulations often comes up as a field of study. In the following study \cite{Mittal2024}, the authors have audited over 100 popular machine learning datasets using their proposed framework that is based on a responsible rubric. Their study highlights vulnerabilities in fairness, privacy, and regulatory compliance. Similarly, transparency requirements both emerging from or targetted to the EU AI Act and the copyright law have been highlighted as important \cite{Buick2024}.

Even though regulatory text is not available in an optimised format for NLP, such models have a key role in meeting regulatory standards, naturally, as they are fit for statistical processing of text. This survey \cite{Jain2025} considers several advancements in natural language processing, along with their limitations. They highlight limitations, such as lack of reliable methods to verify regulatory requirements. Furthermore, domain-specific approaches to bridging the gap between regulatory compliance and standardised evaluation has been proposed for medical device classification \cite{Han2025}. In their study the authors test a multitude of machine learning models, ranging from traditional machine learning to deep learning, in an attempt to discover the most efficient model for that particular task. Besides the comparative analysis, it also demonstrates a benchmark for that particular domain.

The EU AI Act is a milestone regulatory effort. It establishes guidelines, boundaries, and context to evaluate compliance of AI systems. Therefore, the scientific community has been releasing and developing resources that will aid AI development stakeholders to increase their compliance with the act. COMPL-AI \cite{Guldimann2025} is a comprehensive framework that identifies, interprets, and translates broad regulatory requirements into measurable technical requirements. The open-source benchmarking suite has a particular focus on LLMs. A parameterised regulatory learning space has been proposed \cite{Lewis2025} to reduce uncertainties in the enforcement of the EU AI Act, as well as to enable the learning process of stakeholders to interact and apply legal measures. Furthermore, this study \cite{Diaz2025}, presents a technical assessment for challenges that may arise from manufacturers of medical devices when qualifying automated inspections based on deep learning within the context of the EU AI Act. Finally, a RAG system has been proposed as a chatbot for real-time and context-aware compliance verification via retrieval of relevant text.

While the broader literature on regulatory compliance and evaluation of AI models provides valuable insights, a dataset targeted to the fundamental tasks of the AI Act remains limited. To the best of our knowledge this is the first attempt to create a benchmarking dataset to promote the development of natural language processing AI systems to aid with compliance and tangible insight extraction from the EU AI Act.

\section{Background \& Methodology}
\label{sec:methodology}
\subsection{Background}
The establishment of a written legal code resulted in a notable increase in societal stability. The establishment of regulations that transcend memory and personality has been instrumental in fostering consistency over time. This has been achieved by constraining the arbitrariness of writing, thereby ensuring its coherence and predictability. Against this broader historical backdrop, contemporary efforts to govern emerging technologies, most notably artificial intelligence, can be read as a continuation of the same stabilizing impulse through codified, written regulation. The EU AI Act\footnote{Regulation EU 2024/1689, EU Artificial Intelligence Act, European Parliament, 2024} constitutes the first comprehensive legal framework for AI on a global scale. The relevant authorities will enforce it by August 2026. Furthermore, the implementation of the AI Act will vary across chapters, sections, and articles, while the full effect will be realised by 2027. These early actions signalled the EU's intention to regulate AI in a way that balances innovation with the protection of fundamental rights.  

The EU AI Act applies a familiar legal philosophy to a new technological domain. It ranks artificial intelligence by degrees of moral and civic danger, and it calibrates permissible power accordingly. At the top are ``unacceptable risk'' or prohibited systems, deemed a clear threat to safety, livelihoods, or fundamental rights, and therefore prohibited outright through eight bans, including manipulative or deceptive AI, exploitation of vulnerabilities, social scoring, individual criminal-offence risk prediction, indiscriminate scraping to build facial-recognition databases, emotion recognition in workplaces and education, biometric categorisation that infers protected characteristics, and real-time remote biometric identification by law enforcement in public spaces. These prohibitions have applied since February 2025 and are accompanied by the Commission's guidance to support compliance. A second tier designates ``high-risk'' uses across critical infrastructure, education, medical and product safety, employment, access to essential services, biometric applications, law enforcement, migration and border management, and judicial or democratic processes. To deploy these systems, one should satisfy demanding pre-market obligations, including risk management, high-quality datasets to reduce discrimination, activity logging for traceability, extensive technical documentation, clear information for deployers, effective human oversight, and strong standards of accuracy, robustness, and cybersecurity, with rules taking effect in phases from August 2026 to August 2027. Finally, the Act largely exempts ``minimal or no risk'' applications, such as video games or spam filters, implying a principled aim: the law should intervene most strongly where automated systems most directly shape human dignity, equality, and the conditions of accountable public life. Synoptically, 

\begin{equation}
\forall s \, (\text{Banned}(s) \leftrightarrow \exists p \, (\text{Performs}(s, p) \land \text{ProhibitedPractice}(p)))
\end{equation}

\begin{equation}
\begin{split}
\forall s \, (\text{MarketEligible}(s) &\leftrightarrow \\
&((\exists c \, (\text{InCategory}(s, c) \land \text{HighRiskCategory}(c))) \\
&\land \forall o \, (\text{Obligation}(o) \rightarrow \text{Satisfies}(s, o))))
\end{split}
\end{equation}

The first expression captures the AI Act's prohibition rule: an AI system $s$ is banned if and only if it performs at least one practice $p$ that falls within the Act's set of prohibited practices, so exclusion follows precisely from the presence of any unacceptable-risk behaviour. The second expression captures the high-risk regime by linking classification to compliance: a system is market-eligible if and only if it belongs to at least one category $c$ designated as high-risk, and it satisfies every required obligation $o$ attached to high-risk systems. Together, the two formulas represent the Act's core logic. 

\begin{figure}[t]
    \centering
    \includegraphics[width=1.0\textwidth]{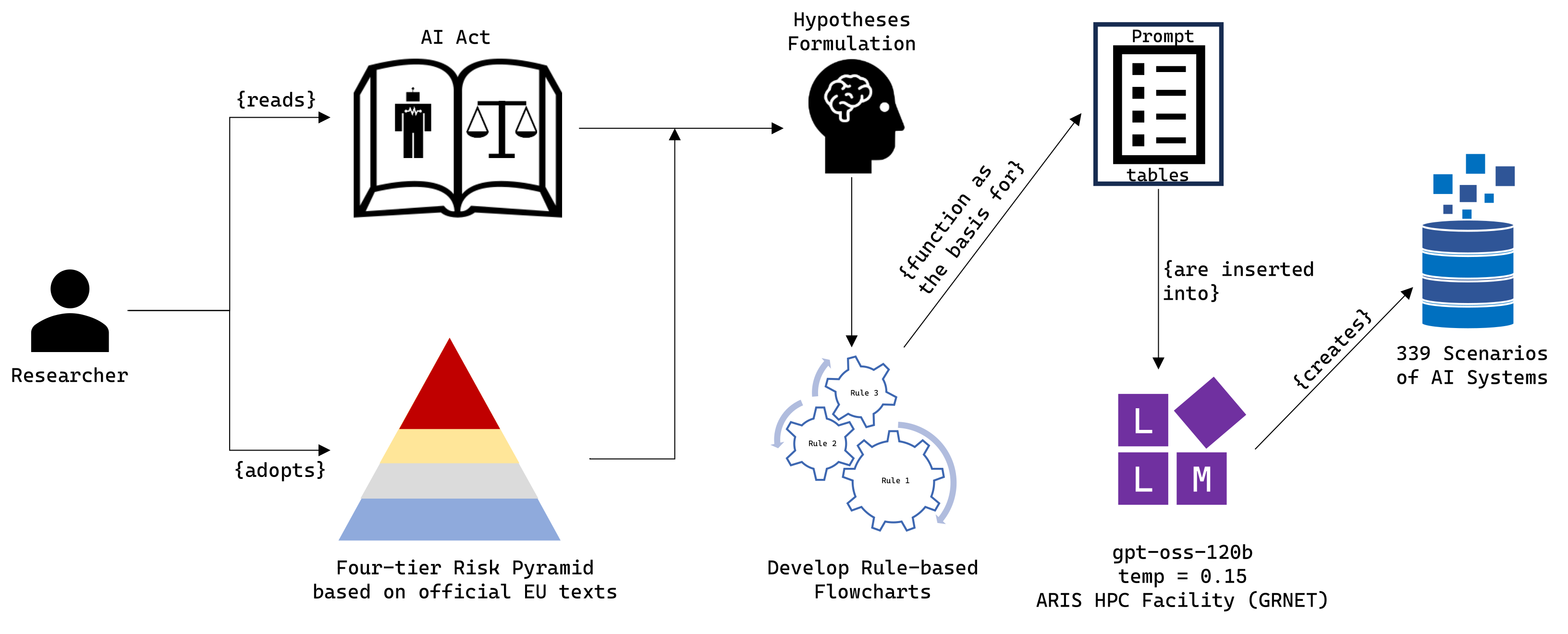}
    \caption{The methodology followed for the creation of AI system use-cases acting as evaluation scenarios in our dataset. Verbs and sentences in brackets represent actions for the connections represented via arrows.}
    \label{fig:method}
\end{figure}

\subsection{Methodology}
In this section, we present the methodology that we followed to create our dataset, along with the hypotheses that were made. Figure \ref{fig:method} presents a visualised version of our methodology. This study adopts a four-tier risk pyramid as the conceptual basis for designing the structure of prompts about AI system scenarios, based on the selected risk classification. We adopt the following categorisation of AI systems: minimal-risk, limited-risk, high-risk, and prohibited systems (unacceptable risk). Although the AI Act's operative provisions do not always use exactly these labels as a taxonomy, EU institutional guidance consistently communicates the AI Act through four risk levels. The European Commission explicitly presents this framing as the Act's risk-based approach\footnote{https://digital-strategy.ec.europa.eu/en/policies/regulatory-framework-ai}. Each tier can serve as a perspicuous ``prompt cosmos'' for dataset generation. The following four basic hypotheses have been utilised to establish a coherent framework for creating AI system scenarios for different risk categories. We present the hypotheses below:

\begin{itemize}
    \item \emph{Hypothesis 1}: The prohibited category of AI systems is directly grounded in the AI Act's binding prohibitions, more specifically in Article 5.  
    \item \emph{Hypothesis 2}: The high-risk category is analogous to the AI systems classified as high-risk under Article 6, which are operationalised via Article 6's classification rules and Annex III. 
    \item \emph{Hypothesis 3}: The limited-risk category of AI systems is grounded in the AI Act's binding articles about transparency and data governance, more specifically in Articles 50 and 10.  
    \item \emph{Hypothesis 4}: The minimal-risk category is treated as a residual bucket. If an AI system is not prohibited (Article 5), not high-risk (Article 6), and does not trigger the specific transparency and data obligations (Articles 50 and 10). 
\end{itemize}

Following the delineation of the four-tier risk hierarchy, the subsequent methodological progression entailed the translation of the AI Act's legal classification logic into explicit, rule-based decision flows. We achieve this step via manual doctrinal reading of the regulation. The purpose was to ensure that the downstream prompt design and dataset generation will rest on human-authored, legally traceable criteria, rather than model-generated interpretations. Specifically, we constructed question-based flowcharts that encode the AI Act's classification triggers as a deterministic sequence of checks. Thus, we generate two primary flow types. The first component is an analysis of prohibited-systems flow, grounded in the prohibitions of Article~5, operationalised as a set of ``does the system do X?'' questions corresponding to the main prohibited practices and their internal conditionalities. The second component is a high-risk identification flow grounded in Article~6 and its interaction with Annex~III, including the AI Act's structured logic. Methodologically, these flowcharts serve as rule elicitation artifacts: they make the AI Act's classification reasoning auditable and operational, as each path yields a category outcome that can be used to condition prompts. We utilise the resulting decision flows  as the basis for the subsequent stages of the study, namely the construction of category-specific prompts and the generation of scenarios that are consistent with the AI Act's rule logic while remaining distinct across the selected risk categories.

Building on the four-tier risk approach and the manually derived rule-based flows, we operationalised each risk tier through a category-specific prompt template for scenario generation. At this stage, we aim to encode the legal classification logic into the prompt structure itself, so that the large language model would be constrained to produce scenarios with the following properties. These properties are: (i) align with the intended risk category, (ii) remain internally consistent with the AI Act's triggers and exceptions, and (iii) are output in a uniform, machine-readable format suitable for dataset construction. In practice, we developed four prompt templates, one for each level, namely: prohibited, high-risk, limited-risk, and minimal-risk. Each of these templates indicate a consistent architectural framework, comprising a system role that assigned the model the stance of a legal analyst of the AI Act, a context field that instructed the insertion of relevant legal provisions, and a task section that required the generation of multiple ``scenario configuration'' under explicit constraints. Across the four templates, the prompts ensure uniform output by requiring a predetermined set of fields. These include, but are not limited to, role in the AI value chain (provider/deployer), intended use, system type, input data, domain, and (where relevant) risk mitigation. Additionally, the prompts impose diversity constraints (e.g., avoiding repeated domains) to reduce redundancy and enhance coverage. Synoptically, the manual doctrinal reading and flowcharting stage produces the human-authored rule logic, while the prompt templates instantiate that logic as ex ante constraints that guide scenario generation through a large language model. Table \ref{tab:prompt} presents the prompt template for prohibited scenario generation. 

\begin{table}[t]
    \caption{Prompt template for generation of prohibited scenarios, ellipsis indicates additional omitted entries.}
    \label{tab:prompt}
    \begin{tabular}{lp{0.72\textwidth}}
    \toprule
    \textbf{Field} & \textbf{Input} \\
    \midrule
      System: & You are a legal analyst specialising in the EU AI Act, with expertise in interpreting Article 5 (Prohibited AI Practices). Your task is to create scenario configurations that explicitly violate the prohibited risk-level of the EU AI Act, based only on the provided context. The output must be a valid JSON file with the following structure and requirements. \\
      User: & \\
      Context: & <Text of Article 5> \\
      Task: & Generate a JSON file containing 100 scenario configurations that actively violate the prohibited risk-level of the EU AI Act, based only on the provided context. \\
      Types of Prohibitions: & Scenarios must violate at least one of the following: \\
      & - Subliminal techniques - Exploitation of vulnerabilities \ldots \\
      Exceptions: & Scenarios must not qualify for any of the following exceptions: \\
      & - Risk assessment of natural persons for criminal purposes (Not prohibited if: The AI system supports the human assessment of a person's involvement in criminal activity, only if the assessment is already based on objective and verifiable facts directly linked to the crime.) \ldots \\
      Requirements: & - Role in the AI value chain (as per EU AI Act): Choose only between provider or deployer. \ldots \\
      Example Format: & \{``role'': ``Provider'', ``intended\_use'': ``AI system that assigns social credit scores to citizens based on their online behaviour, restricting access to public services for those with low scores'',   ``system\_type'': ``Social scoring AI system'', ``input\_data'': ``Biometric data, browsing history, social media activity, geolocation data, facial recognition'', ``domain'': ``Public services''\} \\ 
      Instructions: & - Base all scenarios strictly on the provided context (Article 5 of the EU AI Act). - Each scenario must clearly violate at least one of the prohibited practices under Article 5. \ldots \\
      \bottomrule
  \end{tabular}
\end{table}

We generate the scenarios using \verb$gpt-oss-120b$ because it combines a high parameter count with practical deployability on single-GPU configurations. We achieved single-GPU running after MXFP4 optimisation via utilising the llama.cpp\footnote{https://github.com/ggml-org/llama.cpp} library. Single-GPU run enables parallel execution of multiple generation runs and thereby reducing overall computational cost. This model is distributed under the permissive Apache 2.0 licence, which supports reuse and downstream dissemination of scientific artefacts derived from the model's outputs. We executed the generation pipeline on the ARIS supercomputing infrastructure operated by GRNET S.A., using GPU-accelerated nodes equipped with four NVIDIA A100 GPUs. The resulting dataset contains 339 scenarios (70 prohibited, 86 high-risk, 84 limited, and 99 minimal), linked to an average of 9.8 related articles per scenario (3350 total) and an average of 10.5 actionable compliance/obligation points per scenario (3572 total). In order to minimise creative variance and ensure the reproducibility and factuality of outputs, a low temperature of 0.15 was selected. 

\section{Dataset Overview}
\label{sec:overview}
\subsection{Dataset Structure}
The core motivation behind the design of the dataset is to facilitate the development and evaluation of NLP models, and particular RAG systems, that support alignment with the EU AI Act regulation. Despite, we strive to generalise its applicability and reach to additional use scenarios and roles. The dataset consists of two main files: the scenarios files and the general question-answering file. The main target audience for this dataset are developers of AI systems, however we identify several other audiences. The following audiences could also benefit from our dataset: (i) legal experts looking to improve their decision boundaries regarding risk-level classification, (ii) educators providing clear examples drawn from the regulation itself for educational purposes, (iii) SMEs using practical examples to avoid prohibited practices.

The scenarios file contains scenarios generated following the document grounded process described in the previous section. It contains the following fields: (i) Role: Provider/Deployer, (ii) Intended use of the AI system, (iii) Type of AI system, (iv) Type of data used as input, (v) Domain of operation, (vi) Related articles from the EU AI Act, (vii) Obligations according to the EU AI Act, (viii) Risk level classification according to the EU AI Act. Table \ref{tab:scenarios} shows an example of the dataset.

\begin{table}[t]
    \caption{Entry from the scenarios file of the proposed dataset.}
    \label{tab:scenarios}
    \begin{tabular}{lp{0.8\textwidth}}
    \toprule
    \textbf{Field} & \textbf{Example} \\
    \midrule
    Role & Provider \\
    Intended use & Deploy a hidden‑audio influence platform that subtly nudges shoppers to purchase high‑margin products without their conscious awareness \\
    System type & Subconscious influence engine \\
    Input data & Audio streams, browsing history, purchase records, demographic profiles \\
    Domain & Retail \\
    Related Articles & [5, 10, 16, 27, 50] \\
    Obligations & ["Immediately halt any deployment of the audio influence engine that uses subliminal techniques beyond users\' consciousness, as this practice is prohibited under Article 5(a).", \ldots] \\
    Risk level & Prohibited \\
    \bottomrule
  \end{tabular}
\end{table}

The general QA pairs contains question answer pairs regarding the EU AI Act regulation. It contains fields such as: (i) question, (ii) answer, and (iii) relevant article from where the question and answer pair was drawn. Table \ref{tab:qa_pairs} presents an example of the QA pairs. Both files and the code to reproduce the results can be found in GitHub\footnote{https://github.com/davidath/ai-act-evaluation-benchmark}.

\begin{table}[t]
    \caption{Entry from the QA pairs file of the proposed dataset.}
    \label{tab:qa_pairs}
    \begin{tabular}{lp{0.8\textwidth}}
    \toprule
    \textbf{Field} & \textbf{Example} \\
    \midrule
    Question & What is the primary purpose of the AI Regulation? \\
    Answer & To improve the internal market, promote trustworthy, human‑centric AI, protect health, safety and fundamental rights, and support innovation. \\
    Relevant article & 1 \\
    \bottomrule
  \end{tabular}
\end{table}

\subsection{Edge Cases}
As described in Section \ref{sec:methodology}, we provided to the LLM some relevant articles as context in order to facilitate the generation of scenarios that would cover all the risk levels found in the regulation. In qualitative checks performed on the dataset, the language model remains factual to the articles provided. However, upscaling to yield a diverse set of scenarios, leads to the inclusions of some edge cases. Qualitative checks indicate that this behaviour is more prominent in prohibited and high-risk levels, and less prominent in limited and minimal levels.

On one hand, the process of upscaling necessitates some loosening of the constraints in order to satisfy all requirements. On the other hand, regulations are often somewhat ambiguous. The relevant article using as context for generation of prohibited scenarios, namely Article 5 ``Prohibited AI Practices'' contains unique terms that are not found within the rest of the document. These distinctive terms are of semantic importance for that particular article, while also being highly explicit. For example, the term ``subliminal'' is a frequent term used in Article 5. Regulations aim to create clear and written rules that standardise behaviour, process, or requirements within some specific context and are not particularly effective to act as tutorials or documents to act as decision boundaries. Therefore, the limits of which behaviour can be classified as subliminal or not may not be very direct. For example an AI system that shows ``barely perceptible visual prompts to increase subscription upgrades'', in that particular scenario a barely perceptible visual prompt may be so rapid that the user is not influenced or affected by that prompt. In other words, the boundaries that consist of subliminal behaviour are not clearly indicated by the regulation itself. 

Similarly, in high-risk level scenarios, there are some scenarios that due to ambiguity of the regulation and the language model could either fall under the prohibited or high-risk scenario. For example, ``AI system that manipulates social media content to influence voting behaviour in an election'', on high-level observation this appears to be a high-risk scenario. However, independently voting could be considered as fundamental right and therefore an AI system that is against that premise can be considered as a prohibited practice. While these edge cases may slightly drift from the original text, they facilitate variety and is of particular interest for the validation of retrieval and decision from embedding and language models.

\section{Use case: EU AI Act Risk-Level Classification}
\label{sec:use-case}
As mentioned previously, the main objective of the data generation is to assess, evaluate and refine agents operating on the EU AI Act, and their performance. To test the reasoning capabilities of such agents, we propose the following use-case and classification task. The EU AI Act risk-level classification task has the objective to correctly predict the risk-level of a scenario, with configuration similar to the ones presented in Section \ref{sec:overview}. The respective AI system should classify from the following pool of labels: \emph{prohibited} for systems that make use of unacceptable practices, \emph{high-risk} for systems that could pose significant risk, \emph{limited} for systems that operate in or adjacent to critical domains but do not pose significant harm or systemic risk or their potential for harm can be mitigated by design, context, user control or compliance with relevant obligations, \emph{minimal} for systems that do not pose significant physical, emotional, or economic risks to individuals or society. The following use case and configuration act more as a showcasing rather than an actual evaluation report on the configuration. For that particular reason, we maintain a simplistic and minimal configuration setting and reporting.

To test the reasoning capabilities of a RAG framework we employ the following configuration. We convert the raw text of the AI Articles into embeddings using \verb+jina-embeddings-v3+\footnote{https://huggingface.co/jinaai/jina-embeddings-v3} which is a BERT-based embedding model to create a vector database. The vector database is utilised for retrieval of the relevant AI Act articles via comparison and matching, after transformation of the query in the databases embedding space using the same embedding model. A specialised empirically tested rubric generated via facilitation from a medium-sized model from Mistral (\verb+mistral-small3.2+\footnote{https://ollama.com/library/mistral-small3.2}) is utilised for query rewriting. We employ Spotify\'s Annoy algorithm \footnote{https://github.com/spotify/annoy} which is a GPU enabled approximate nearest neighbour implementation. Lastly, we prompt the mistral-small3.2 to generate a response to the query, given the relevant articles as context. The prompt is customised for the purposes of standardised output. According to \cite{Gao2023}, this is classified as an advanced RAG configuration, with pre-retrieval techniques. 

We utilise the following pair of query and prompt, ``\emph{Classify the risk category (prohibited, high-risk, limited, minimal) for an AI system with the following specifications: - Role: \texttt{\$role} - Domain: \texttt{\$domain} - Type: \texttt{\$system\_type} - Input data: \texttt{\$input\_data} - Intended use: \texttt{\$intended\_use}. Provide a detailed rationale based on ethical, legal, and safety considerations, addressing: privacy concerns, accuracy and reliability, potential misuse, compliance with legal standards impact on civil liberties balance between security needs and individual rights.}'' and ``\emph{Task: Answer the following question: \texttt{\$query} Output: The output should be a single word representing the level of the system. Do not provide any additional information''}.

\begin{table}[t]
    \caption{Detailed report on the performance of an advanced RAG system using the proposed dataset on the task of risk-level classification.}
    \label{tab:risk-performance}
    \centering
    \begin{tabular}{cccc}
    \toprule
    \textbf{Class} & \textbf{Precision} & \textbf{Recall} & \textbf{F1-Score} \\
    \midrule
    prohibited & 0.86 & 0.89 & 0.87 \\
    high-risk & 0.82 & 0.88 & 0.85 \\
    limited & 0.51 & 0.88 & 0.65 \\
    minimal & 0.97 & 0.29 & 0.45 \\
    weighted average & 0.79 & 0.71 & 0.69 \\
    \bottomrule
  \end{tabular}
\end{table}

In Table \ref{tab:risk-performance} the performance of the above configuration is presented. The overall performance of the configuration suggests a fairly solid precision (0.79) and recall (0.71), with a moderate balance between the two (0.69). The proposed configuration achieves good performance on the prohibited and high-risk classes, which can be considered as quite critical for the success of the classifier. The F1-score for prohibited (0.87) and high-risk (0.85) indicate that there is a clear decision boundary for these class. This is quite expected as the limits, practices, and operation setting of these cases are explicitly stated within the EU AI Act. The unacceptable practices are meticulously defined within Article 5 (Prohibited AI practices) in order to be avoided by all roles and stakeholders involved in development of AI systems. At the same time, due to high-risk AI systems operating in critical domains their operation is allowed under special conditions and thus are also defined in Article 6 (Classification Rules for High-Risk AI Systems). 

On the other hand, limited and minimal scenarios are not clearly defined within the EU AI Act, while their decision boundaries can be reached by reversing the logic for addressing prohibited and high-risk cases, distincting between the two is not straightforward and frequently be made in either direction depending on the selection of keywords. This is also highlighted by the performance of the configuration, limited scenarios achieve moderate precision (0.53) with moderately-high (0.88) recall resulting in moderate f1-score (0.65) indicating the existence of an abundance of false positives and therefore overprediction of that particular class. At the same time, minimal scenarios present a very high precision (0.97) with quite low recall (0.29) resulting in a bit lower than moderate performance (0.45), indicating that this class is underpredicted. The above results are quite expected due to context provided from the articles of the AI Act not being sufficient for prediction of these classes. To improve the performance for these classes a subset of techniques should be implemented. Techniques such as, adjustment of query rewriting, query and prompting using manually extracted criteria, using Chain-Of-Thought, using models with larger capacity or employing a more fine-grained chunking strategy, however employing these techniques is outside the scope of this work.

\section{Conclusions \& Future Work}
\label{sec:conc}
In this paper, we present the creation and generation of a dataset that aims to address evaluation and validation of NLP models that facilitate regulatory compliance with the EU AI Act. We introduce our methodology and the processing steps, which reflects the text extraction of the raw document in a transparent, open, and reproducible manner. Furthermore, we showcase a use case along with experimentation and results for a RAG system operating on the EU AI Act to produce risk-level classification. This showcase demonstrates the dataset's usefulness in evaluation of the AI models, and acting as a baseline for comparison.

Besides, presenting the overview and use case of the dataset, this paper produces a formalisation of machine learning tasks that can be extracted from the EU AI Act. With this formalisation, we aim to motivate a democratised development of the AI models to facilitate the legal aspect of Trustworthy AI. Through this methodology, we additionally aim to address other roles and stakeholders related to our studies, such as legal experts, educators or SMEs, by disseminating knowledge extracted via statistical means which highly reflect the original document's core structure and logic. Through the above, we aim to lower the introduction boundary and facilitate discussion on addressing the shortcomings of the EU AI Act (e.g., clear definition of limiter or minimal risk-levels), along with propagation of good practices and general guidelines.

The work presented in this paper aims to mainly address the data scarcity when coming to evaluating AI models operating on the EU AI Act. To this end, we orchestrate and delineate several prompts based on empirical rubrics made from humanly extracted rules and knowledge. Our methodology aims to employ techniques to ensure faithful and document-grounded output via technical means (e.g., large context lengths, low temperature, prompt instructions on remaining faithful to context). Besides technical means, we try to employ to the best of our resources to a qualitative human validation of the entries generated. However, even though we try to safeguard our approach to the best of our capabilities, it remains a work based on statistical models. Processing via statistical models should be addressed as such and should not replace actual legal advice. Therefore, validation with legal experts or other practitioners in the field of compliance with European regulation is required. However, to address the initial issues for data scarcity, this validation should take place in a framework that does not require constant supervision from the expert in order to reduce manual work required, e.g., a semi-supervised framework.

Furthermore, the development and release of open source language models is a constant procedure that increasingly produces better and better models. Experimentation with such models is a future direction, however to abide to reproducibility and transparency it should be models that allow their repurpose into such causes such as the Apache 2.0 or other similar licences.  

Besides, refining the technical means of processing such as the choice of LLMs and refinement of prompts, the regulatory landscape is constantly evolving along with the AI development. In practice, this means that corrections and additional material to the EU AI Act are consistently created and disseminated by the official channels. Involving such material to address the shortcomings of the original document, but also extending to other aspects of trustworthy AI such as including ethical principles or robust technical guidelines is a prominent future direction.
 
\paragraph{Acknowledgements} 
This work has been funded by the Digital Europe Programme (DIGITAL) under grant agreement No. 101146490 — DIGITAL-2022-CLOUD-AI-B-03. The inference of large language models was supported by computational time granted from the National Infrastructures for Research and Technology S.A. (GRNET) in the National HPC facility — ARIS — under project ID 018047 (pr018047\_gpu).
\bibliography{main}

\end{document}